\documentclass[runningheads]{llncs}
\usepackage{graphicx}
\usepackage{comment}
\usepackage{color}
\usepackage{ftnxtra}
\usepackage{fnpos}
\usepackage{amsmath}
\usepackage{subfigure}
\usepackage{multirow}
\usepackage{hyphenat}

\begin{document}
\pagestyle{headings}
\mainmatter

\title{A Multi-modal Machine Learning Approach and Toolkit to Automate Recognition of Early Stages of Dementia among British Sign Language Users}



\titlerunning{Toolkit to Automate Recognition of Early Stages of Dementia in BSL Users}
%
\author{Xing Liang\inst{1}\and
Anastassia Angelopoulou\inst{2}\and
Epaminondas Kapetanios\inst{2}\and Bencie Woll\inst{3}\and Reda Al-batat\inst{2}\and Tyron Woolfe\inst{3}
}
\authorrunning{X.Liang et al.}
%
\institute{IoT and Security Research Group, University of Greenwich, UK
\email{\{x.liang\}@greenwich.ac.uk}
\and
Cognitive Computing Research Lab, University of Westminster, UK
\email{\{agelopa,kapetae,r.albatat\}@westminster.ac.uk} \and
Deafness Cognition and Language Research Centre, University College London, UK 
\email{\{b.woll,twoolfe\}@ucl.ac.uk}}

\maketitle

\begin{abstract}
 The ageing population trend is correlated with an increased prevalence of acquired cognitive impairments such as dementia. Although there is no cure for dementia, a timely diagnosis helps in obtaining necessary support and appropriate medication. Researchers are working urgently to develop effective technological tools that can help doctors undertake early identification of cognitive disorder. In particular, screening for dementia in ageing Deaf signers of British Sign Language (BSL) poses additional challenges as the diagnostic process is bound up with conditions such as quality and availability of interpreters, as well as appropriate questionnaires and cognitive tests. On the other hand, deep learning based approaches for image and video analysis and understanding are promising, particularly the adoption of Convolutional Neural Network (CNN), which require large amounts of training data. In this paper, however, we demonstrate novelty in the following way: a) a multi-modal machine learning based automatic recognition toolkit for early stages of dementia among BSL users in that features from several parts of the body contributing to the sign envelope, e.g., hand-arm movements and facial expressions, are combined, b) universality in that it is possible to apply our technique to users of any sign language, since it is language independent, c) given the trade-off between complexity and accuracy of machine learning (ML) prediction models as well as the limited amount of training and testing data being available, we show that our approach is not over-fitted and has the potential to scale up.

\end{abstract}

\section{Introduction}
British Sign Language (BSL) is a natural human language, which, like other sign languages, uses movements of the hands, body and face for linguistic expression. Recognising dementia in the signers of BSL,  however, is still an open research field, since there is very little information available about dementia in this population. This is also exacerbated by the fact that there are few clinicians with appropriate communication skills and experience working with BSL users. Diagnosis of dementia is subject to the quality of cognitive tests and BSL interpreters alike. Hence, the Deaf community currently receives unequal access to diagnosis and care for acquired neurological impairments, with consequent poorer outcomes and increased care costs \cite{Atkinson2001}. 

Facing this challenge, we outlined a machine learning based methodological approach and developed a toolkit capable of automatically recognising early stages of dementia without the need for sign language translation or interpretation. Our approach and tool were inspired by the following two key cross-disciplinary knowledge contributors:

a) Recent clinical observations suggesting that there may be differences between signers with dementia and healthy signers with regards to the envelope of sign space (sign trajectories/depth/speed) and expressions of the face. These clinical observations indicate that signers who have dementia use restricted sign space and limited facial expression compared to healthy deaf controls. In this context, we did not focus only on the hand movements, but also on other features from the BSL user's body, e.g., facial expressions.

b) Recent advances in machine learning based approaches spearheaded by CNN, also known as the {\it Deep Learning} approach. These, however, cannot be applied without taking into consideration contextual restrictions such as availability of large amounts of training datasets, and lack of real world test data. We introduce a deep learning based sub-network 
for feature extraction together with the CNN approach for diagnostic classification, which yields better performance and is a good alternative to handle limited data.

In this context, we proposed a multi-featured machine learning methodological approach paving the way to the development of a toolkit. The promising results for its application towards screening for dementia among BSL users lie with using features other than those bound to overt cognitive testing by using language translation and interpretation. 
Our methodological approach comprises several stages. The first stage of research focuses on analysing the motion patterns of the sign space envelope in terms of sign trajectory and sign speed by deploying a real-time hand movement trajectory tracking model~\cite{liang2019} based on OpenPose\footnote{https://github.com/CMU-Perceptual-Computing-Lab/openpose}$^{,}$\footnote{https://github.com/ildoonet/tf-pose-estimation} library. The second stage involves the extraction of the facial expressions of deaf signers by deploying a real-time facial analysis model based on dlib library\footnote{http://dlib.net/} to identify active and non-active facial expressions.  The third stage is to trace elbow joint distribution based on OpenPose library, taken as an additional feature related to the sign space envelope. Based on the differences in patterns obtained from facial and trajectory motion data, the further stage of research implements both VGG16 \cite{Simonyan2015} and ResNet-50 \cite{He2016} networks using transfer learning from image recognition tasks to incrementally identify and improve recognition rates for Mild Cognitive Impairment (MCI) (i.e. pre-dementia).  Performance evaluation of the research work is based on datasets available from the Deafness Cognition and Language Research Centre (DCAL) at UCL, which has a range of video recordings of over 500 signers who have volunteered to participate in research. It should be noted that as the deaf BSL-using population is estimated to be around 50,000, the size of this database is equivalent to 1\% of the deaf population. Figure \ref{The Proposed Pipeline for Dementia Screening} shows the pipeline and high-level overview of the network design. The main contributions of this paper are as follows:

\begin{figure}
    \begin{center}
        \includegraphics[width=0.95\textwidth]{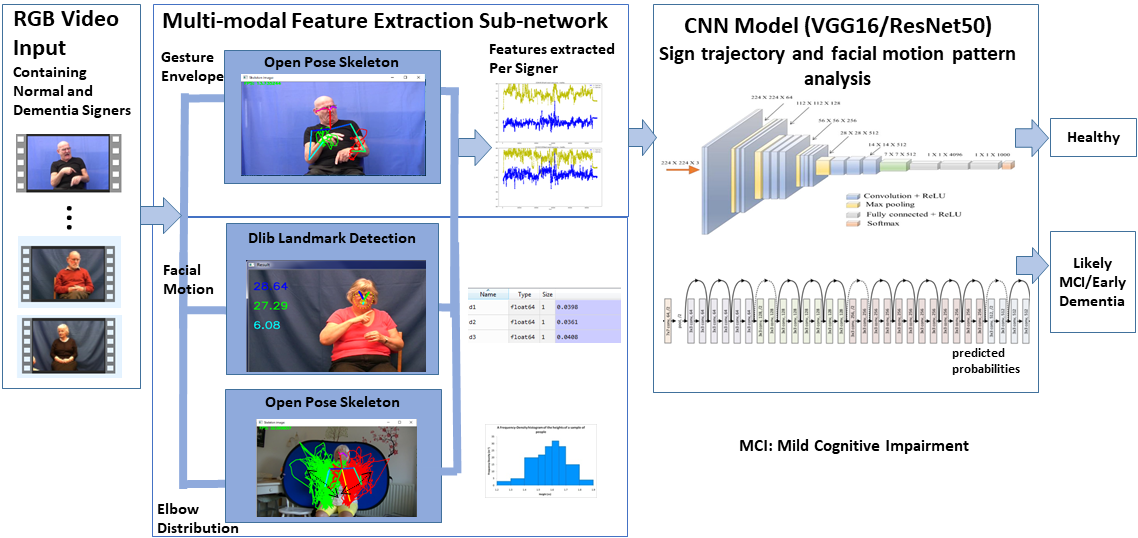}
        \caption{The Proposed Pipeline for Dementia Screening}
        \label{The Proposed Pipeline for Dementia Screening}
    \end{center}
\end{figure}

\begin{enumerate}
\item We outline a methodology for the prelimnary identification of early stage dementia among BSL users based on sign language independent features such as: 
  \subitem an accurate and robust real-time hand trajectory tracking model, in which both {\it sign trajectory} to extract sign space envelope and {\it{sign  speed}}  to identify acquired neurological impairment associated with motor symptoms are tracked.
  \subitem a real-time facial analysis model that can identify and differentiate {\it   active} and {\it non-active facial expressions} of a signer.
 \subitem an elbow distribution model that can identify the {\it motion characteristics of the elbow joint} during signing. 
 \item We present an automated screening toolkit for early stage dementia assessment with good test set performance of 87.88\% in accuracy, 0.93 in ROC,  0.87 in F1 Score for positive MCI/dementia screening results. 
 As the proposed system uses normal 2D videos without requiring any ICT/medical facilities setup, it is economical, simple, flexible, and adaptable.
\end{enumerate}
 The paper is structured as follows: Section \ref{s1} gives an overview of the related work. Section \ref{s2} outlines the methodological approach followed by section \ref{s3} with the discussion of experimental design and results. A conclusion provides a summary of the key contributions and results of this paper.

\section{Related Work}\label{s1}
Recent advances in computer vision and greater availability in medical imaging with improved quality have increased the opportunities to develop deep learning approaches for automated detection and quantification of diseases, such as Alzheimer and dementia \cite{Pellegrini2018}. Many of these techniques have been applied to the classification of MR imaging, CT scan imaging, FDG-PET scan imaging or the combined imaging of above, by comparing MCI patients to healthy controls, to distinguish different types or stages of MCI and accelerated features of ageing  \cite{Young2018,Spasova2019,Lu2018,Huang2019}. Jo et al. in \cite{Jo2019} reviewed the deep learning papers on Alzheimer (published between January 2013 and July 2018) with the conclusion that four of the studies used combination of deep learning and traditional machine learning approaches, and twelve used deep learning approaches. Due to currently limited dataset, we also found that  ensemble  the deep learning approaches for diagnostic classification with the traditional machine  learning  methods  for  feature  extraction  yielded  a better  performance. 

In terms of dementia diagnosis \cite{Astell2019}, there have been increasing applications of various machine learning approaches, most commonly with imaging data for diagnosis and disease progression \cite{Negin2018,Dallora2017,Iizuka2019} and less frequently in non-imaging studies focused on demographic data, cognitive measures \cite{Bhagyashree2018}, and unobtrusive monitoring of gait patterns over time \cite{Dodge2012}. In \cite{Dodge2012}, walking speed and its daily variability may be an early marker of the development of MCI. These and other real-time measures of function may offer novel ways of detecting transition phases leading to dementia, which could be another potential research extension to our toolkit, since the real-time hand trajectory tracking sub-model has the potential to track a patient's daily walking pattern and pose recognition as well. AVEID, an interesting method introduced in \cite{Parekh2018}, uses an automatic video system for measuring engagement in dementia, focusing on behaviour on observational scales and emotion detection. AVEID focused on passive engagement on gaze and emotion detection, while our method focuses on sign and facial motion analysis in active signing conversation.   

\section{Methodology}\label{s2}
In this paper, we present a multi-modal feature extraction sub-network inspired by practical clinical needs, together with the experimental findings associated with the sub-network. Each feature extraction model is discussed in greater detail in the following sub-sections and for each method we assume that the subjects are in front of the camera with only the face, upper body, and arms visible. The input to the system is short-term clipped videos.
Different extracted motion features will be fed into the CNN network to classify a BSL signer as healthy or atypical. We present the first phase work on automatic assessment of early stage dementia based on real-time hand movement trajectory motion patterns and focusing on performance comparisons between the VGG16 and ResNet-50 networks. Performance evaluation of the research work is based on datasets available from the BSL Corpus\footnote{British Sign Language Corpus Project  https://bslcorpusproject.org/.} at DCAL UCL, a collection of 2D video clips of 250 Deaf signers of BSL from 8 regions of the UK; and two additional datasets: a set of data collected for a previous funded project\footnote{Overcoming obstacles to the early identification of dementia in the signing Deaf community},
and a set of signer data collected for the present study.

\subsection{Dataset}
From the video recordings, we selected 40 case studies of signers (20M, 20F) aged between 60 and 90 years; 21 are signers considered to be healthy cases based on the British Sign Language Cognitive Screen (BSL-CS); 9 are signers identified as having Mild Cognitive Impairment (MCI) on the basis of the BSL-CS; and 10 are signers diagnosed with mild MCI through clinical assessment. We consider those 19 cases as MCI (i.e. early dementia) cases, either identified through the BSL-CS or clinically. Balanced datasets (21 Healthy, 19 MCI) are created in order to decrease the risk of leading to a falsely perceived positive effect of accuracy due to the bias towards one class. While this number may appear small, it represents around 2\% of the population of signers likely to have MCI, based on its prevalence in the UK. 
As the video clip for each case is about 20 minutes in length, we segmented each into 4-5 short video clips - 4 minutes in length - and fed the segmented short video clip to the multi-modal feature extraction sub-network. The feasibility study and experimental findings discussed in Section \ref{secExperimentFindings} show that the segmented video clips represent the characteristics of individual signers. In this way, we were able to increase the size of the dataset from 40 to 162 clips. Of the 162, 79 have MCI, and  83 are cognitively healthy. 

\subsection{Real-time Hand Trajectory Tracking Model}
\begin{figure}
    \begin{center}
        \includegraphics[width=0.7\textwidth]{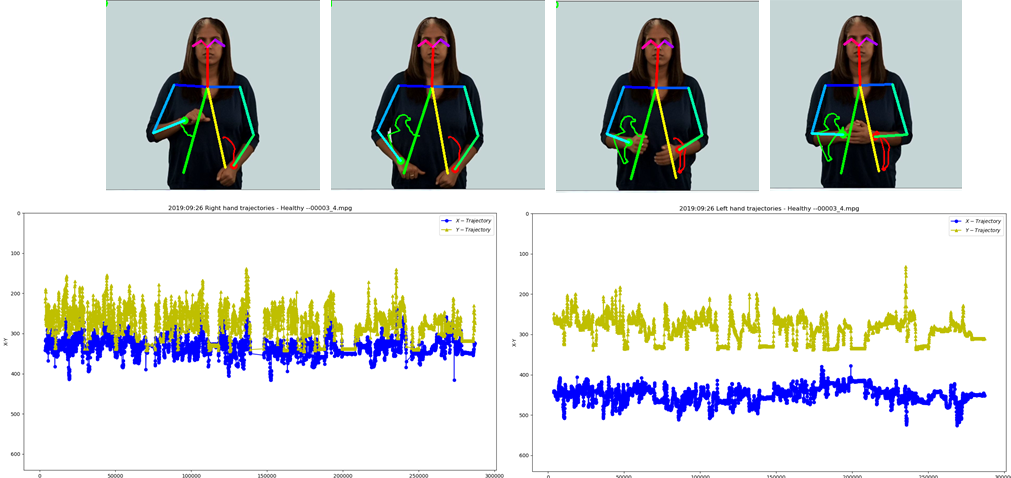}
        \caption{Real-time Hand Trajectory Tracking (top) and 2D Left- and Right- Hand Trajectory (bottom)}
        \label{Trajectorycombine}
    \end{center}
\end{figure}

OpenPose, developed by Carnegie Mellon University, is one of the state-of-the-art methods for human pose estimation, processing images through a 2-branch multi-stage CNN~\cite{Cao2017}. The real-time hand movement trajectory tracking model is developed 
based on the OpenPose Mobilenet Thin model \cite{OpenPoseTensorflow}. A detailed evaluation of  tracking performance is discussed in~\cite{liang2019}. The inputs to the system are brief clipped videos, 
and only 14 upper body parts in the image are outputted from the tracking model. These are: eyes, nose, ears, neck, shoulders, elbows, wrists, and hips. The hand movement trajectory is obtained via wrist joint motion trajectories. The curve of the hand movement trajectory is connected by the location of the wrist joint keypoints to track left- and right-hand limb movements across sequential video frames in a rapid and unique way. Figure \ref{Trajectorycombine} (top), demonstrates the tracking process for the sign FARM. 

Figure \ref{Trajectorycombine} (bottom) is the left- and right-hand trajectories obtained from the tracking model plotted by wrist location X and Y coordinates over time in a 2D plot. 
It shows how hand motion changes over time, which gives a clear indication of hand movement speed (X-axis speed based on 2D coordinate changes, and Y-axis speed based on 2D coordinate changes). A spiky trajectory indicates more changes within a shorter period, thus faster hand movement. Hand movement speed patterns can be easily identified to analyse acquired neurological impairments associated with motor symptoms (i.e. slower movement), as in Parkinson’s disease. 

\subsection{Real-time Facial Analysis Model}
\begin{figure}
    \begin{center}
        \includegraphics[width=0.6\textwidth]{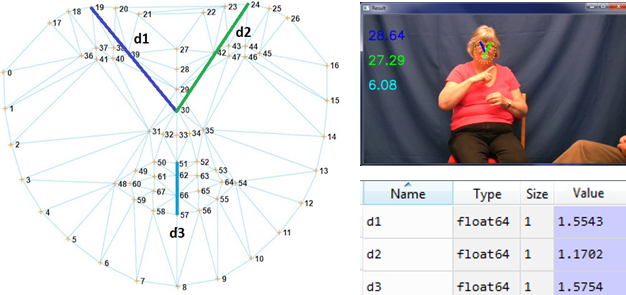}
        \caption{Facial Motion Tracking of a Signer}
        \label{FacailMotion}
    \end{center}
\end{figure}

The facial analysis model was implemented based on a facial landmark detector inside the Dlib library, in order to analyse a signer’s facial expressions \cite{Kazemi2014}. The face detector uses the classic Histogram of Oriented Gradients (HOG) feature combined with a linear classifier, an image pyramid, and a sliding window detection scheme. The pre-trained facial landmark detector is used to estimate the location of 68 (x, y) coordinates that map to facial features (Figure~\ref{FacailMotion}).

\begin{figure}
    \begin{center}
        \includegraphics[width=0.55\textwidth]{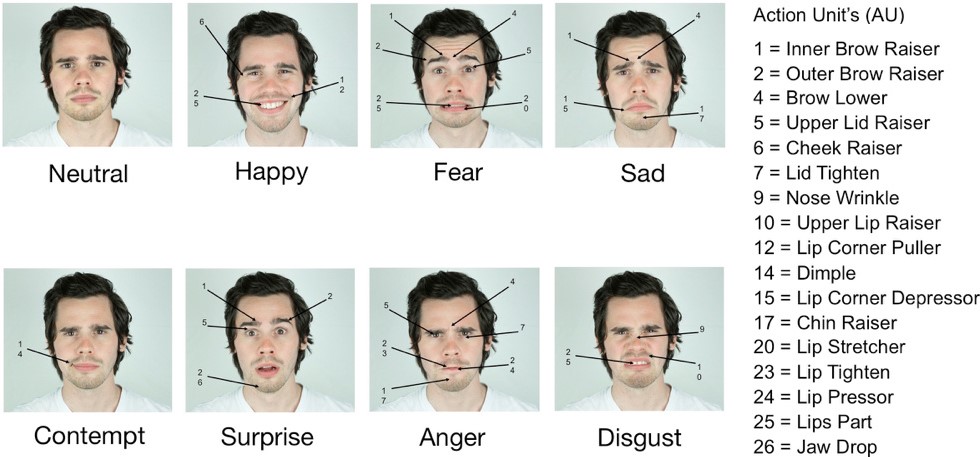}
        \caption{Common Facial Expressions}
        \label{CommonFacial}
    \end{center}
\end{figure}

As shown in Figure \ref{CommonFacial}\footnote{https://www.eiagroup.com/knowledge/facial-expressions/}, earlier psychological research~\cite{charles1998} identified seven universal common facial expressions: Happiness, Sadness, Fear, Disgust, Anger, Contempt and Surprise. Facial muscle movements for these expressions include lips and brows (Figure \ref{CommonFacial}). Therefore, the facial analysis model was implemented for the purpose of extract subtle facial muscle movement by calculating the average Euclidean distance differences between the nose and right brow as d1, nose and left brow as d2, and upper and lower lips as d3 for a given signer over a sequence of video frames (Figure \ref{FacailMotion}). The vector [d1, d2, d3] is an indicator of a signer’s facial expression and is used to classify a signer as having an active or non-active facial expression. 
\begin{equation}
     d1, d2, d3 = \frac{\displaystyle\sum\limits_{t=1}^T \arrowvert d^{t+1} -d^{t} \arrowvert}{T}
\end{equation} 
    where T = Total number of frames that facial landmarks are detected.
    
\subsection{Elbow Distribution Model}
\begin{figure}
    \begin{center}
        \includegraphics[width=0.55\textwidth]{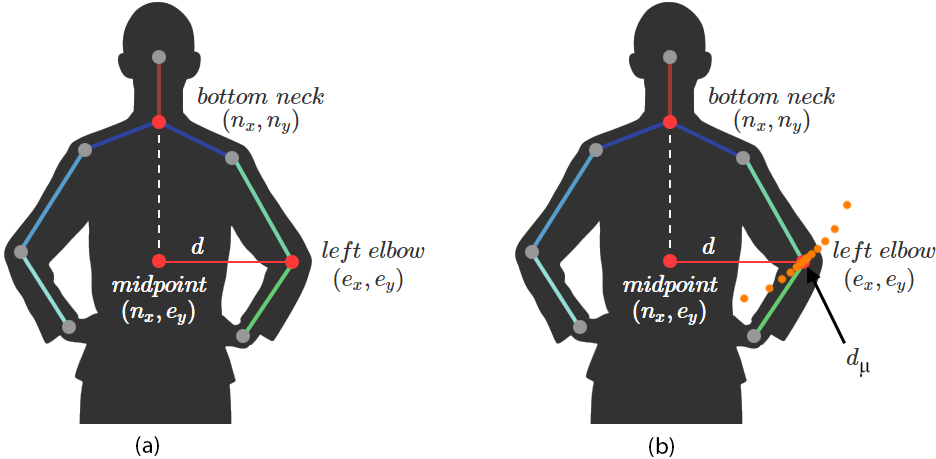}
        \caption{(a) Elbow tracking distance from the midpoint. (b) Shifted coordinate with mean distance calculated}
        \label{elbow_tracking_distance}
    \end{center}
\end{figure}

The elbow distribution model extracts and represents the motion characteristics of elbow joint movement during signing, based on OpenPose upper body keypoints. The Euclidean distance $d$ is calculated between the elbow joint coordinate and a relative midpoint of the body in a given frame. This is illustrated in Figure~\ref{elbow_tracking_distance}(a), where the midpoint location on the frame is made up of the x-coordinate of the neck and the y-coordinate of the elbow joint.
If $J_{e,n}^t$ represents distances of joints elbow and neck ($e$,$n$) at time $t$, such as $J_{e,n}^t = [X_{e,n}^t , Y_{e,n}^t]$ then $d$ calculates the distance descriptor:
\begin{equation}
     {\displaystyle d ={\sqrt {(X_{n}^t-X_{e}^t)^{2}+(Y_{n}^t-Y_{e}^t)^{2}}}}
\end{equation} 
for each frame, resulting in $N$ distances $d$, where $N$ is the number of frames. In order to get a distribution representation of elbow motion, a virtual coordinate origin is created, which is the mean distance calculated as $
 \displaystyle d_{\mu} = \frac{\sum_i^N \textbf{d}}{N}$,  which can be seen as the resting position of the elbow. Then a relative distance is calculated from this origin $d_{\mu}$ to the elbow joint for each frame, resulting in the many distances shown in Figure~\ref{elbow_tracking_distance}(b) as orange dots.
If the relative distance is $<0$ it is closer to the body than the resting distance, and if it is $>0$, it is further away. This is a much better representation of elbow joint movement as it distinguishes between near and far elbow motion. These points can be represented by a histogram which can then be fed into the CNN model as an additional feature.

\subsection{CNN Models}
In this section, we summarise the architecture of the VGG16 and ResNet-50 
implemented for the early dementia classification, focusing on data pre-processing, architecture overview, and transfer learning in model training. 

\subsubsection{Data Pre-processing}
Prior to classification, we first vertical stack a signer's left-hand trajectory image over the associated right-hand trajectory image obtained from the real-time hand trajectory tracking model, and label the 162 stacked input trajectory images as pairs \begin{equation}
 (X, Y)= \lbrace (X_{1}, Y_{1}), ..., (X_{i}, Y_{i}), ..., (X_{N}, Y_{N})\rbrace \: \: \: \: \:   (N = 162) 
\end{equation}
where $  X_{i}$   is the i-th observation (image dimension: 1400 $\times$ 1558 $\times$ 3) from the MCI and Healthy datasets. The classification has the corresponding class label  $  Yi \in \lbrace 0, 1\rbrace$, with early MCI (Dementia) as class 0 and Healthy as class 1. 
The input images are further normalized by subtracting the ImageNet data mean and changed the input shape dimensions to 224$\times$224$\times$3 to be ready for the Keras deep learning CNN networks. 

\subsubsection{VGG16 and ResNet-50 Architecture}
\begin{figure}
    \begin{center}
        \includegraphics[width=0.90\textwidth]{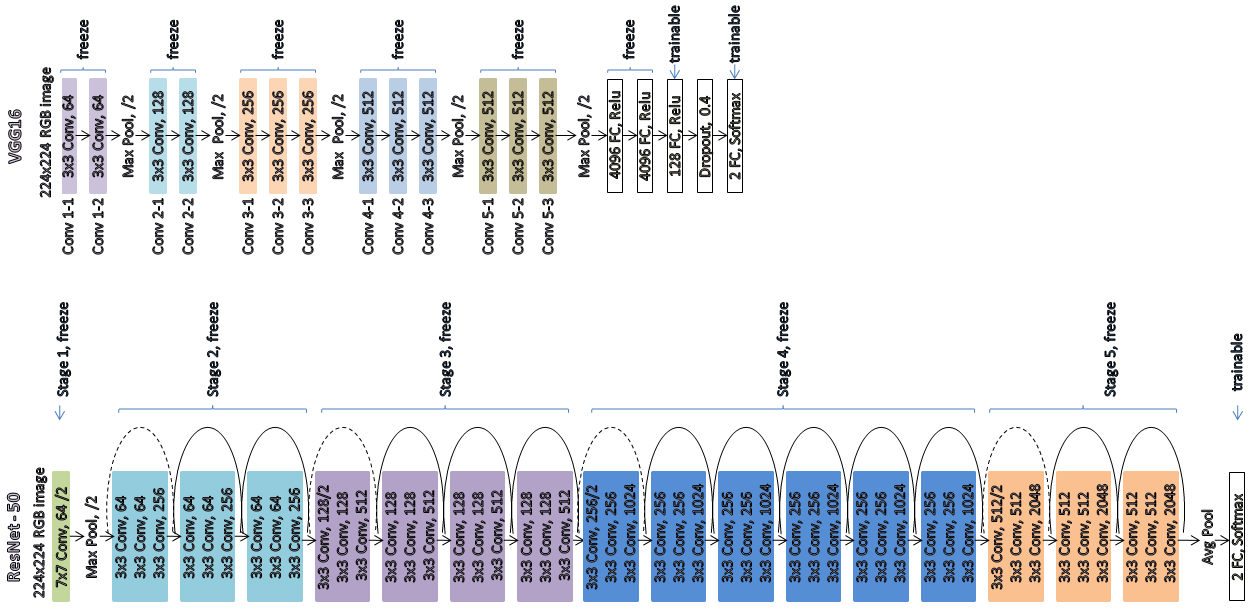}
        \caption{VGG16 and ResNet-50 Architecture}
        \label{vgg16resnet50}
    \end{center}
\end{figure}

In our approach, we have used VGG16 and ResNet-50 as the base models with transfer learning  to transfer the parameters pre-trained for 1000 object detection task on ImageNet dataset to recognise hand movement trajectory images for early MCI screening. Figure \ref{vgg16resnet50} shows the network architecture that we implemented by fine tuning VGG16 and training ResNet-50 as a classifier alone.  

1) VGG16 Architecture: The VGG16 network \cite{Simonyan2015} with 13 convolutional and 3 fully connected (FC) layers, i.e. 16 trainable weight layers, were the basis of the Visual Geometry Group (VGG) submission to the ImageNet Challenge 2014, achieving 92.7\% top-5 test accuracy, and securing first and second places in the classification and localization track respectively.  Due to the very small dataset, we fine tune the VGG 16 network by freezing the Convolutional (Covn) layers and two Fully Connected (FC) layers, and only retrain the last two layers, with 524,674 parameters trainable in total (see Figure \ref{vgg16resnet50}). Subsequently, a softmax layer for binary classification is applied to discriminate the two labels: Healthy and  MCI, producing two numerical values of which the sum becomes 1.0.

Several strategies are used to combat overfitting. A dropout layer is implemented after the last FC \cite{Srivastava2014}, randomly dropping 40\% of the units and their connections during training. An intuitive explanation of its efficacy is that each unit learns to extract useful features on its own with different sets of randomly chosen inputs. As a result, each hidden unit is more robust to random fluctuations and learns a generally useful transformation. Moreover, EarlyStopping is used to halt the training of the network at the right time to avoid overfitting. EarlyStopping  callback is configured to monitor the loss on the validation dataset with the patience argument set to 15. The training process is stopped after 15 epochs when there is no improvement on the validation dataset. 

2) ResNet-50 Architecture: Residual Networks (ResNets) \cite{He2016} introduce skip connections to skip blocks of convolutional layers, forming a residual block. These stacked residual blocks greatly improve training efficiency and largely resolve the vanishing gradient problem present in deep networks. This model won the ImageNet challenge in 2015; the top 5 accuracy for ResNet-50 is 93.29\%.
As complex models with many parameters are more prone to overfitting with a small dataset, we train ResNet-50 as a classifier alone rather than fine tune it (see Figure \ref{vgg16resnet50}). Only a softmax layer for binary classification is applied, which introduces 4098 trainable parameters. EarlyStopping callback is also configured to halt the training of the network in order to avoid overfitting.

\section{Experiments and Analysis}\label{s3}
\subsection{Implementation}
The networks mentioned above were constructed using Python 3.6.8, OpenCV 3.4.2, and Tensorflow 1.12. VGG16 and ResNet-50 were built with the Keras deep learning library \cite{chollet2015}, using Tensorflow as backend. We employed a Windows desktop with two Nvidia GeForce GTX 1080Ti adapter cards and 3.3 GHz Intel Core i9-7900X CPU with 16 GB RAM. During training, dropout was deployed in fully connected layers and EarlyStopping was used to avoid overfitting. To accelerate the training process and avoid local minimums, we used Adam algorithm with its default parameter setting (learning rate=0.001, beta 1=0.9, beta 2=0.999) as the training optimizer \cite{Kingma2015}. Batch size was set to 3 when training VGG16 network and 1 when training ResNet-50 network, as small mini-batch sizes provide more up-to-date gradient calculations and yield more stable and reliable training \cite{YoshuaBengio2012,Master2015}. In training it took several ms per epoch, with ResNet-50 quicker than the other because of less in training parameters. As an ordinary training schedule contains 100 epochs, in most cases, the training loss would converge in 40 epochs for VGG16 and 5 epochs for ResNet-50. During training, the parameters of the networks were saved via Keras callbacks to monitor EarlyStopping to save the best weights. These parameters were used to run the test and validation sets later. During test and validation, accuracies and Receiver Operating Characteristic (ROC) curves of the classification were calculated, and the network with the highest accuracy and area under ROC was chosen as the final classifier.

\subsection{Results and Discussion}

\subsubsection{Experiment Findings}\label{secExperimentFindings}
\begin{figure}
    \begin{center}
        \includegraphics[width=0.9\textwidth]{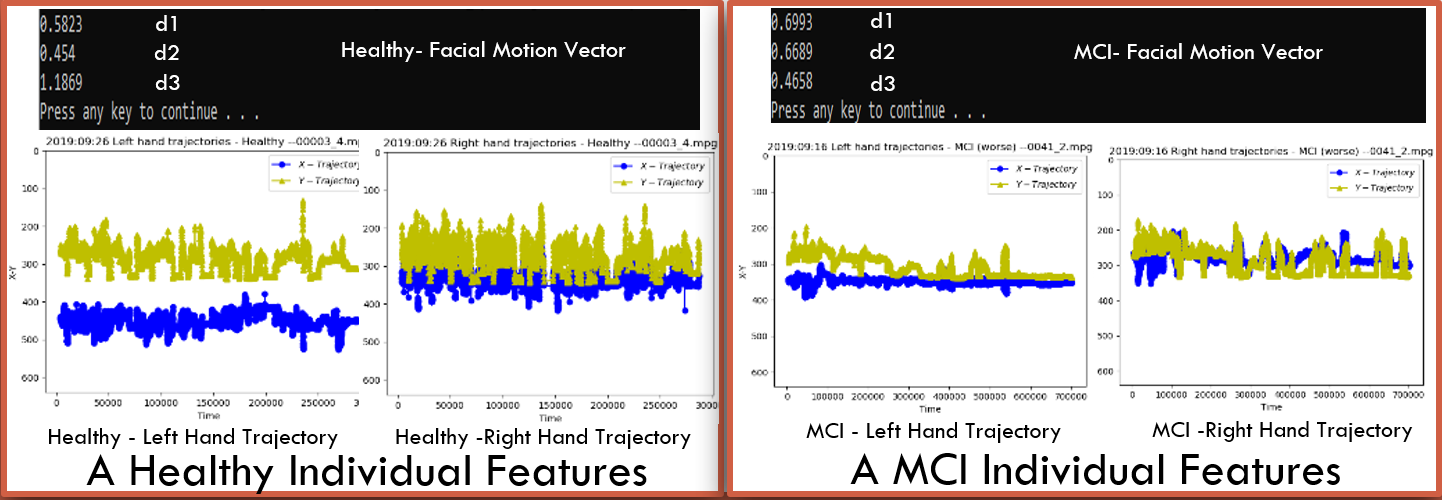}
        \caption{Experiment Finding}
        \label{ExperimentFinding}
    \end{center}
\end{figure}

\begin{figure}
    \begin{center}
        \includegraphics[width=0.8\textwidth]{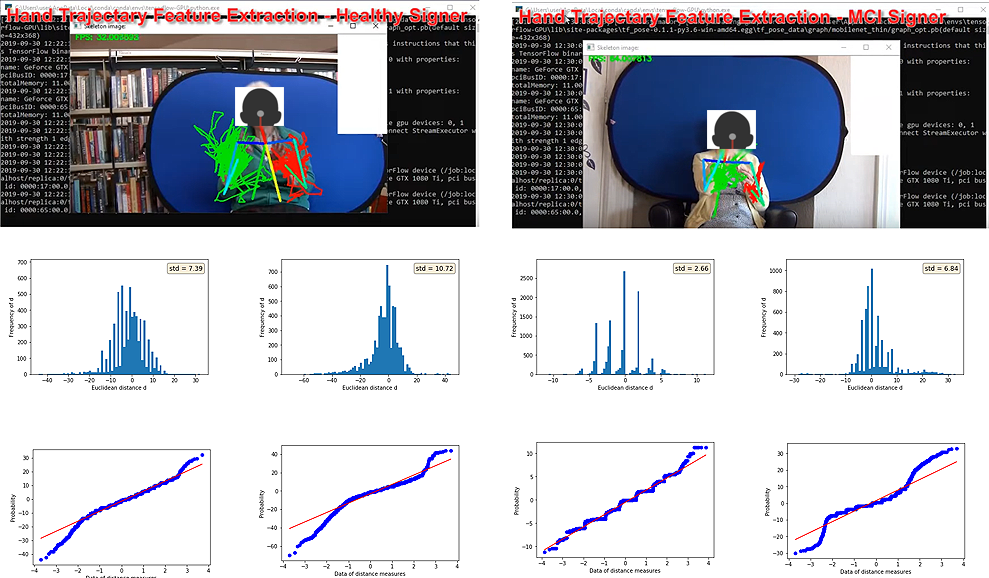}
        \caption{The top row shows signing space for a healthy (left) and an MCI (right) signer. The bottom row shows the acquired histograms and normal probability plots for both hands. For data protection purposes both faces have been covered.}
        \label{elbow_results}
    \end{center}
\end{figure}

In Figure \ref{ExperimentFinding}, feature extraction results show that in a greater number of cases a signer with MCI produces a sign trajectory that resembles a straight line rather than the spiky trajectory characteristic of a healthy signer. In other words, signers with MCI produced more static poses/pauses during signing, with a reduced sign space envelope as indicated by smaller amplitude differences between the top and bottom peaks of the X, Y trajectory lines. At the same time, the Euclidean distance d3 of healthy signers is larger than that of MCI signers, indicating active facial movements by healthy signers. This proves the clinical observation concept of differences between signers with MCI and healthy signers in the envelope of sign space and face movements, with the former using smaller sign space and limited facial expression.
In addition to space and facial expression, the elbow distribution model demonstrates restricted movement around the elbow axis with a lower standard deviation and a skewed distribution for the MCI signer compared to the healthy signer where the distribution is normal (Figure \ref{elbow_results}). 
\subsubsection{Performance Evaluation}
\begin{figure}
    
     \begin{center}
        \subfigure{%
           \label{Fig_cf_test_vgg16_without_norm}
           \includegraphics[width=0.22\textwidth]{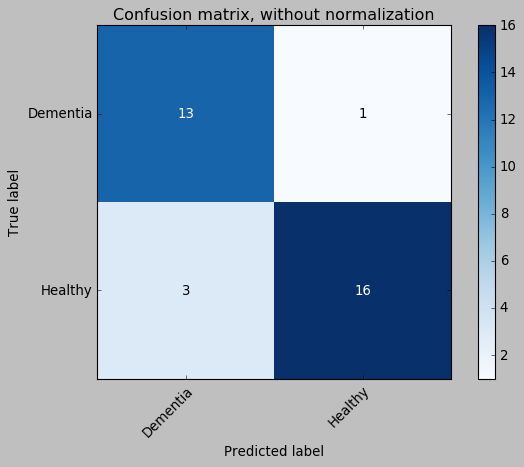}
        } %
        \subfigure{%
           \label{Fig_cf_test_vgg16_norm}
           \includegraphics[width=0.22\textwidth]{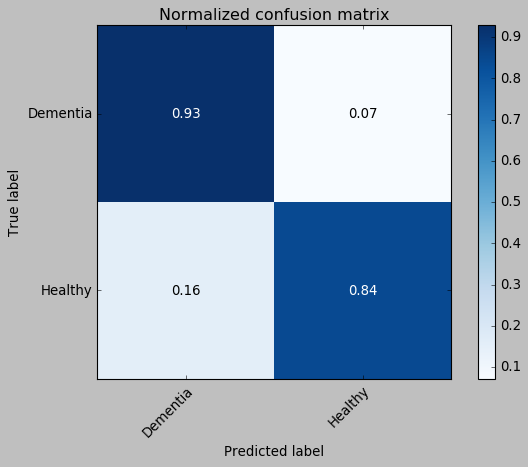}
        } %
        \subfigure{%
            \label{Fig_cf_test_resnet50_without_norm}
            \includegraphics[width=0.22\textwidth]{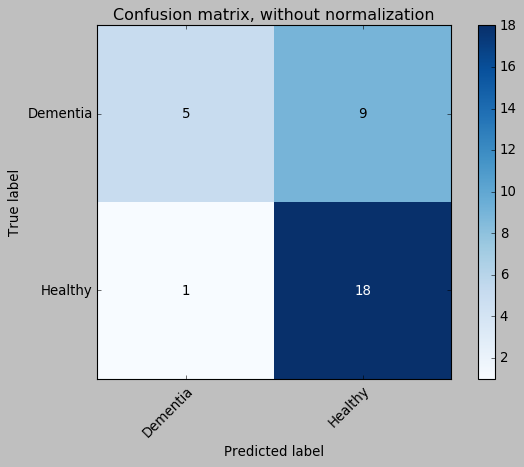}
        }%
        \subfigure{%
        \label{Fig_cf_test_resnet50_norm}
        \includegraphics[width=0.22\textwidth]{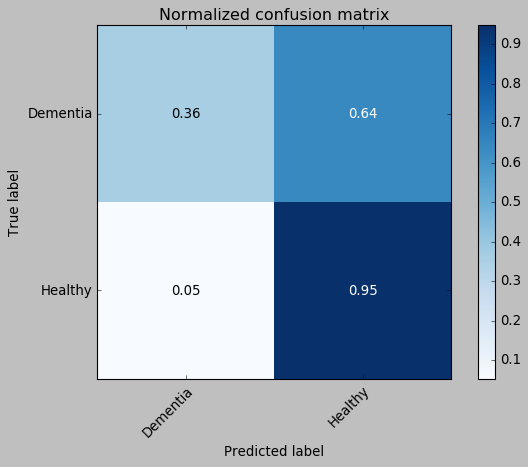}
        }%
    \end{center}
    \caption{%
        Test Set Confusion Matrix of VGG16 (left two) and ResNet-50 (right two)
     }%
   \label{FigTestConfusionMatrix}
\end{figure}

\begin{figure}
    \begin{center}
        \includegraphics[width=0.35\textwidth]{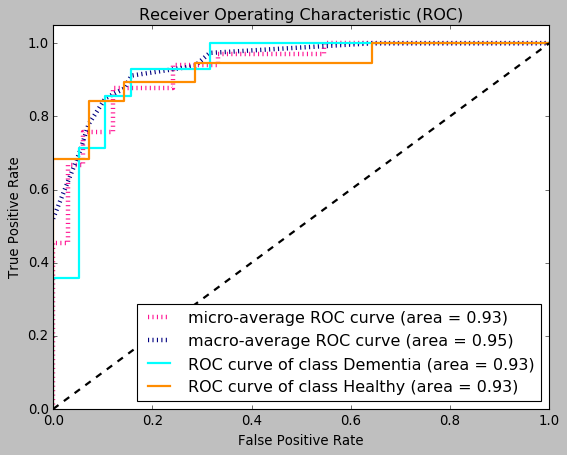}
        \includegraphics[width=0.35\textwidth]{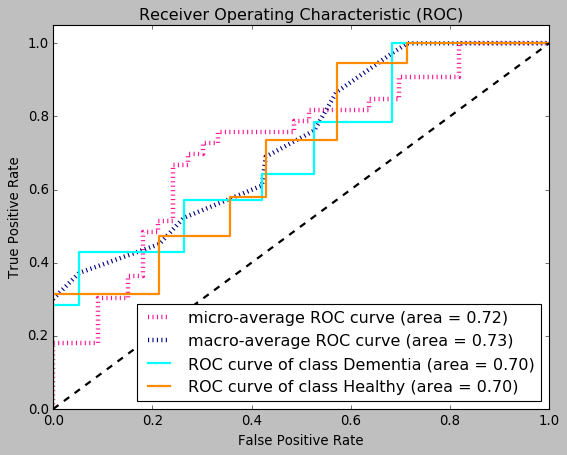}
        \caption{Test Set ROC of VGG16 (left) and ResNet-50 (right)}
        \label{Fig_ROC_test}
    \end{center}
\end{figure}

\begin{figure}
    \begin{center}
        \includegraphics[width=0.35\textwidth]{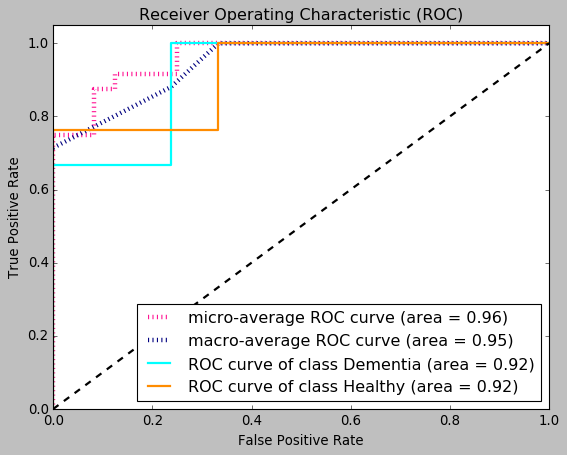}
        \caption{Validation Set ROC on VGG16}
        \label{Fig_ROC_validation}
    \end{center}
\end{figure}

In this section, we have performed a comparative study of VGG16 and ResNet-50 networks. Videos of 40 participants have been segmented into short clips with 162 segmented cases in the training processes. Those segmented samples are randomly partitioned into two subsets with splitting into 80\% for the training set and 20\% for the test set. To validate the model performance, we also kept 6 cases separate (1 MCI and 5 healthy signers) that have not been used in the training process, segmented into 24 cases for performance validation.  The  validation samples is skewed as a result of limited in MCI samples but richer in health samples. More MCI samples are kept in the training/test processes than in the validation. Table \ref{tab:performance-table} shows effectiveness results over 46 participants from different networks. 
\begin{table}
\centering
\caption{Performance Evaluation over VGG16 and RestNet-50 for early MCI screening}
\label{tab:performance-table}
\begin{tabular}{|c|c|c|c|c|c|}
\hline
\multirow{5}{4em}{\textbf{Method}} &
  \multicolumn{3}{c|}{\begin{tabular}[c]{@{}c@{}}40 Participants\\ 21 Healthy, 19 Early MCI\end{tabular}} &
  \multicolumn{2}{c|}{\begin{tabular}[c]{@{}c@{}}6 Participants\\  5 Healthy, 1 Early MCI\end{tabular}} \\ \cline{2-6} 
 &
  \begin{tabular}[c]{@{}c@{}}Train Result \\ (129 segmented cases)\end{tabular} &
  \multicolumn{2}{c|}{\begin{tabular}[c]{@{}c@{}}Test Result\\ (33 segmented cases)\end{tabular}} &
  \multicolumn{2}{c|}{\begin{tabular}[c]{@{}c@{}}Validation Result\\ (24 segmented cases)\end{tabular}} \\ \cline{2-6} 
                   & ACC       & ACC                & ROC           & ACC             & ROC           \\ \hline
\textbf{VGG 16}    & 87.5969\% & \textbf{87.8788\%} & \textbf{0.93} & \textbf{87.5\%} & \textbf{0.96} \\ \hline
\textbf{ResNet-50} & 69.7674\% & 69.6970\%          & 0.72          & 66.6667\%       & 0.73     \\ \hline     
\end{tabular}
\end{table}
The ROC curves are further illustrated in Figure \ref{FigTestConfusionMatrix} and Figure \ref{Fig_ROC_test} based on test set performance. The best performance metrics are achieved by VGG16 with accuracy of 87.8788\%; a micro ROC of 0.93; F1 score for MCI: 0.87, for Healthy: 0.89. Therefore, VGG16 was selected as the baseline classifier and validation was further performed on 24 sub-cases from 6 participants.  
Table 2 summarises validation performance over the baseline classifier  VGG16, and its ROC in Figure \ref{Fig_ROC_validation}. In Table \ref{tab:validation-table}, there are two false positive and one false negative based on sub-case prediction, but the model has a correct high confidence prediction rate on most of the sub-cases. If prediction confidence is averaged  over all of the sub-cases from a participant, and predict the result, the model achieved 100\% accuracy in validation performance. 

\begin{table}
\centering
\caption{Validation Performance over Baseline Classifier - VGG16}
\label{tab:validation-table}
\scalebox{0.78}{
\begin{tabular}{ccccccc}
\hline
\multicolumn{1}{|c|}{\multirow{1}{6em}{\textbf{Participant No}}} &
  \multicolumn{1}{c|}{\multirow{1}{5em}{\textbf{Sub-case}}} &
  \multicolumn{2}{c|}{\textbf{\begin{tabular}[c]{@{}c@{}}Prediction\\ Confidence\end{tabular}}} &
  \multicolumn{1}{c|}{\multirow{1}{9.5em}{\textbf{\begin{tabular}[c]{@{}c@{}}Prediction Result \\ based on Sub-case\end{tabular}}}} &
  \multicolumn{1}{c|}{\multirow{1}{10.5em}{\textbf{\begin{tabular}[c]{@{}c@{}}Prediction Result\\ based on Participant\end{tabular}}}} &
  \multicolumn{1}{c|}{\multirow{1}{4em}{\textbf{Ground Truth}}} \\ \cline{3-4}
\multicolumn{1}{|c|}{} &
  \multicolumn{1}{c|}{} &
  \multicolumn{1}{c|}{\textbf{MCI}} &
  \multicolumn{1}{c|}{\textbf{Health}} &
  \multicolumn{1}{c|}{} &
  \multicolumn{1}{c|}{} &
  \multicolumn{1}{c|}{} \\ \hline
\multicolumn{1}{|c|}{\multirow{1}{4em}{1}} &
  \multicolumn{1}{c|}{1\_1} &
  \multicolumn{1}{c|}{0.63} &
  \multicolumn{1}{c|}{0.37} &
  \multicolumn{1}{c|}{MCI} &
  \multicolumn{1}{c|}{\multirow{1}{4em}{Healthy}} &
  \multicolumn{1}{c|}{\multirow{1}{4em}{Healthy}} \\ \cline{2-5}
\multicolumn{1}{|c|}{} &
  \multicolumn{1}{c|}{1\_2} &
  \multicolumn{1}{c|}{0.43} &
  \multicolumn{1}{c|}{0.57} &
  \multicolumn{1}{c|}{Healthy} &
  \multicolumn{1}{c|}{} &
  \multicolumn{1}{c|}{} \\ \cline{2-5}
\multicolumn{1}{|c|}{} &
  \multicolumn{1}{c|}{1\_3} &
  \multicolumn{1}{c|}{0.39} &
  \multicolumn{1}{c|}{0.61} &
  \multicolumn{1}{c|}{Healthy} &
  \multicolumn{1}{c|}{} &
  \multicolumn{1}{c|}{} \\ \cline{2-5}
\multicolumn{1}{|c|}{} &
  \multicolumn{1}{c|}{1\_4} &
  \multicolumn{1}{c|}{0.27} &
  \multicolumn{1}{c|}{0.73} &
  \multicolumn{1}{c|}{Healthy} &
  \multicolumn{1}{c|}{} &
  \multicolumn{1}{c|}{} \\ \cline{2-5}
\multicolumn{1}{|c|}{} &
  \multicolumn{1}{c|}{1\_5} &
  \multicolumn{1}{c|}{0.40} &
  \multicolumn{1}{c|}{0.60} &
  \multicolumn{1}{c|}{Healthy} &
  \multicolumn{1}{c|}{} &
  \multicolumn{1}{c|}{} \\ \hline
\multicolumn{1}{|c|}{\multirow{1}{4em}{2}} &
  \multicolumn{1}{c|}{2\_1} &
  \multicolumn{1}{c|}{0.13} &
  \multicolumn{1}{c|}{0.87} &
  \multicolumn{1}{c|}{Healthy} &
  \multicolumn{1}{c|}{\multirow{1}{4em}{Healthy}} &
  \multicolumn{1}{c|}{\multirow{1}{4em}{Healthy}} \\ \cline{2-5}
\multicolumn{1}{|c|}{} &
  \multicolumn{1}{c|}{2\_2} &
  \multicolumn{1}{c|}{0.02} &
  \multicolumn{1}{c|}{0.98} &
  \multicolumn{1}{c|}{Healthy} &
  \multicolumn{1}{c|}{} &
  \multicolumn{1}{c|}{} \\ \cline{2-5}
\multicolumn{1}{|c|}{} &
  \multicolumn{1}{c|}{2\_3} &
  \multicolumn{1}{c|}{0.56} &
  \multicolumn{1}{c|}{0.44} &
  \multicolumn{1}{c|}{MCI} &
  \multicolumn{1}{c|}{} &
  \multicolumn{1}{c|}{} \\ \cline{2-5}
\multicolumn{1}{|c|}{} &
  \multicolumn{1}{c|}{2\_4} &
  \multicolumn{1}{c|}{0.23} &
  \multicolumn{1}{c|}{0.77} &
  \multicolumn{1}{c|}{Healthy} &
  \multicolumn{1}{c|}{} &
  \multicolumn{1}{c|}{} \\ \hline
\multicolumn{1}{|c|}{\multirow{1}{4em}{3}} &
  \multicolumn{1}{c|}{3\_1} &
  \multicolumn{1}{c|}{0.08} &
  \multicolumn{1}{c|}{0.92} &
  \multicolumn{1}{c|}{Healthy} &
  \multicolumn{1}{c|}{\multirow{1}{4em}{Healthy}} &
  \multicolumn{1}{c|}{\multirow{1}{4em}{Healthy}} \\ \cline{2-5}
\multicolumn{1}{|c|}{} &
  \multicolumn{1}{c|}{3\_2} &
  \multicolumn{1}{c|}{0.02} &
  \multicolumn{1}{c|}{0.98} &
  \multicolumn{1}{c|}{Healthy} &
  \multicolumn{1}{c|}{} &
  \multicolumn{1}{c|}{} \\ \cline{2-5}
\multicolumn{1}{|c|}{} &
  \multicolumn{1}{c|}{3\_3} &
  \multicolumn{1}{c|}{0.02} &
  \multicolumn{1}{c|}{0.98} &
  \multicolumn{1}{c|}{Healthy} &
  \multicolumn{1}{c|}{} &
  \multicolumn{1}{c|}{} \\ \cline{2-5}
\multicolumn{1}{|c|}{} &
  \multicolumn{1}{c|}{3\_4} &
  \multicolumn{1}{c|}{0.01} &
  \multicolumn{1}{c|}{0.99} &
  \multicolumn{1}{c|}{Healthy} &
  \multicolumn{1}{c|}{} &
  \multicolumn{1}{c|}{} \\ \hline
\multicolumn{1}{|c|}{\multirow{1}{4em}{4}} &
  \multicolumn{1}{c|}{4\_1} &
  \multicolumn{1}{c|}{0.09} &
  \multicolumn{1}{c|}{0.91} &
  \multicolumn{1}{c|}{Healthy} &
  \multicolumn{1}{c|}{\multirow{1}{4em}{Healthy}} &
  \multicolumn{1}{c|}{\multirow{1}{4em}{Healthy}} \\ \cline{2-5}
\multicolumn{1}{|c|}{} &
  \multicolumn{1}{c|}{4\_2} &
  \multicolumn{1}{c|}{0.24} &
  \multicolumn{1}{c|}{0.76} &
  \multicolumn{1}{c|}{Healthy} &
  \multicolumn{1}{c|}{} &
  \multicolumn{1}{c|}{} \\ \cline{2-5}
\multicolumn{1}{|c|}{} &
  \multicolumn{1}{c|}{4\_3} &
  \multicolumn{1}{c|}{0.16} &
  \multicolumn{1}{c|}{0.84} &
  \multicolumn{1}{c|}{Healthy} &
  \multicolumn{1}{c|}{} &
  \multicolumn{1}{c|}{} \\ \cline{2-5}
\multicolumn{1}{|c|}{} &
  \multicolumn{1}{c|}{4\_4} &
  \multicolumn{1}{c|}{0.07} &
  \multicolumn{1}{c|}{0.93} &
  \multicolumn{1}{c|}{Healthy} &
  \multicolumn{1}{c|}{} &
  \multicolumn{1}{c|}{} \\ \hline
\multicolumn{1}{|c|}{\multirow{1}{4em}{5}} &
  \multicolumn{1}{c|}{5\_1} &
  \multicolumn{1}{c|}{0.01} &
  \multicolumn{1}{c|}{0.99} &
  \multicolumn{1}{c|}{Healthy} &
  \multicolumn{1}{c|}{\multirow{1}{4em}{Healthy}} &
  \multicolumn{1}{c|}{\multirow{1}{4em}{Healthy}} \\ \cline{2-5}
\multicolumn{1}{|c|}{} &
  \multicolumn{1}{c|}{5\_2} &
  \multicolumn{1}{c|}{0.01} &
  \multicolumn{1}{c|}{0.99} &
  \multicolumn{1}{c|}{Healthy} &
  \multicolumn{1}{c|}{} &
  \multicolumn{1}{c|}{} \\ \cline{2-5}
\multicolumn{1}{|c|}{} &
  \multicolumn{1}{c|}{5\_3} &
  \multicolumn{1}{c|}{0.00} &
  \multicolumn{1}{c|}{1.00} &
  \multicolumn{1}{c|}{Healthy} &
  \multicolumn{1}{c|}{} &
  \multicolumn{1}{c|}{} \\ \cline{2-5}
\multicolumn{1}{|c|}{} &
  \multicolumn{1}{c|}{5\_4} &
  \multicolumn{1}{c|}{0.07} &
  \multicolumn{1}{c|}{0.93} &
  \multicolumn{1}{c|}{Healthy} &
  \multicolumn{1}{c|}{} &
  \multicolumn{1}{c|}{} \\ \hline
  \multicolumn{1}{|c|}{\multirow{1}{4em}{6}} &
 \multicolumn{1}{c|}{6\_1} &
  \multicolumn{1}{c|}{0.93} &
  \multicolumn{1}{c|}{0.07} &
  \multicolumn{1}{c|}{MCI} &
  \multicolumn{1}{c|}{\multirow{1}{4em}{MCI}} &
  \multicolumn{1}{c|}{\multirow{1}{4em}{MCI}} \\ \cline{2-5}
\multicolumn{1}{|c|}{} &
  \multicolumn{1}{c|}{6\_2} &
  \multicolumn{1}{c|}{0.29} &
  \multicolumn{1}{c|}{0.71} &
  \multicolumn{1}{c|}{Healthy} &
  \multicolumn{1}{c|}{} &
  \multicolumn{1}{c|}{} \\ \cline{2-5}
\multicolumn{1}{|c|}{} &
  \multicolumn{1}{c|}{6\_3} &
  \multicolumn{1}{c|}{0.91} &
  \multicolumn{1}{c|}{0.09} &
  \multicolumn{1}{c|}{MCI} &
  \multicolumn{1}{c|}{} &
  \multicolumn{1}{c|}{} \\ \hline
\end{tabular}}
\end{table}
Furthermore, since a deep learning network can easily become over-fitted with relatively small datasets, comparison against simpler approaches such as logistic regression and SVM is also performed. As stated in \cite{Dreiseitl2002}, logistic regression and artificial neural networks are the models of choice in many medical data classification tasks, with one layer of hidden neurons generally sufficient for classifying most datasets. Therefore, we evaluate our datasets on a 2-layer shallow neural network with 80 neurons in hidden layer and logistic sigmoid activation as its output layer.
\begin{table}
\centering
\caption{Comparing Deep Neural Network Architecture over Shallow Networks  }
\label{tab:deep-shallow-table}
\scalebox{0.99}{
\begin{tabular}{|l|l|l|ll}
\cline{1-3}
                 & Train Accuracy (\%) & Test Accuracy (\%) &  &  \\ \cline{1-3}
VGG16            & 87.5969            & 87.8788           &  &  \\ \cline{1-3}
Shallow Logistic & 86.4865            & 86.1538           &  &  \\ \cline{1-3}
SVM              & 86.8725            & 73.8461           &  &  \\ \cline{1-3}
\end{tabular}
}
\end{table}

Our observations on comparison results in respect with accuracy between shallow (Logistic, SVM) and deep learning CNN prediction models, presented in Table \ref{tab:deep-shallow-table}, show that, for smaller datasets, shallow models are a considerable alternative to deep learning models, since no significant improvement could be shown. Deep learning models, however, have the potential to perform better in the presence of larger datasets \cite{Schindler2016}. Since we aspire to train and apply our model with increasingly larger amounts of data made available, our approach is well justified. The comparisons also highlighted that our ML prediction model is not over-fitted despite the fact that small amounts of training and testing data were available.

\section{Conclusions}
We have outlined a multi-modal machine learning methodological approach and developed a toolkit for an automatic dementia screening system. The toolkit uses VGG16, while focusing on analysing features from various body parts, e.g., facial expressions, comprising the sign space envelope of BSL users recorded in normal 2D videos. As part of our methodology, we report the experimental findings for the multi-modal feature extractor sub-network in terms of hand sign trajectory, facial motion, and elbow distribution, together with performance comparisons between different CNN models in ResNet-50 and VGG16. The experiments show the effectiveness of our machine learning based approach for early stage dementia screening. The results are validated against cognitive assessment scores with a test set performance of 87.88\%, and a validation set performance of 87.5\% over sub-cases, and 100\% over participants. Due to its key features of being economic, simple, flexible, and adaptable, the proposed methodological approach and the implemented toolkit have the potential for use with other sign languages, as well as in screening for other acquired neurological impairments associated with motor changes, such as stroke and Parkinson’s disease in both hearing and deaf people. 

\section*{Acknowledgements}
This work has been supported by The Dunhill Medical Trust grant number RPGF1802$\backslash37$, UK.

\clearpage
%
%
\bibliographystyle{splncs04}
\bibliography{ECCV_2020_ATDA-BSL}

\end{document}